\documentclass{scrartcl}

\usepackage[american]{babel}
\usepackage{natbib}
\usepackage[colorlinks=true, citecolor=blue]{hyperref}
\usepackage{mathtools}
\usepackage{booktabs}
\usepackage{tikz}
\usepackage{amsfonts, amsmath, amssymb}
\usepackage{multirow}
\usepackage{float}
\usepackage{algorithm}
\usepackage{algorithmic}

\newcommand{\x}{\mathbf{x}}
\newcommand{\w}{\mathbf{w}}
\newcommand{\Hh}{\mathcal{H}}
\newcommand{\tf}{\tilde{f}}
\newcommand{\tg}{\tilde{g}}
\newcommand{\bstheta}{\boldsymbol{\theta}}
\newcommand{\bbeta}{\boldsymbol{\beta}}
\DeclareMathOperator*{\argmin}{arg\,min}
\DeclareMathOperator*{\argmax}{arg\,max}

\title{Bayesian Optimization Augmented with Actively Elicited Expert Knowledge}

\author{
Daolang Huang\,$^1$ \hspace{0.5em} Louis Filstroff\,$^1$ \hspace{0.5em} Petrus Mikkola\,$^1$ \\
Runkai Zheng\,$^3$ \hspace{0.5em} Samuel Kaski\,$^{1,2}$ \\
\small $^1$ Department of Computer Science, Aalto University, Finland \\
\small $^2$ Department of Computer Science, University of Manchester, UK \\
\small $^3$ School of Data Science, The Chinese University of Hong Kong, Shenzhen, China
}

\begin{document}

\maketitle

\begin{abstract}
Bayesian optimization (BO) is a well-established method to optimize black-box functions whose direct evaluations are costly. In this paper, we tackle the problem of incorporating expert knowledge into BO, with the goal of further accelerating the optimization, which has received very little attention so far. We design a multi-task learning architecture for this task, with the goal of jointly eliciting the expert knowledge and minimizing the objective function. In particular, this allows for the expert knowledge to be transferred into the BO task. We introduce a specific architecture based on Siamese neural networks to handle the knowledge elicitation from pairwise queries. Experiments on various benchmark functions with both simulated and actual human experts show that the proposed method significantly speeds up BO even when the expert knowledge is biased compared to the objective function.
\end{abstract}

\section{Introduction}\label{sec:intro}

Bayesian optimization (BO) \citep{Jones1998,brochu2010tutorial} has become a well-established class of methods to optimize black-box functions, with applications in hyper-parameter tuning \citep{snoek2012practical}, chemistry \citep{hase2018phoenics}, and material science \citep{zhang2020bayesian}, to cite only a few. Formally, let $f: \mathcal{X} \rightarrow \mathbb{R}$ be a black-box function defined over some compact space $\mathcal{X}$. We assume that evaluating $f$ at some point $\x$ is possible, but expensive. The goal is to find the global optimum $\x^{\star}$, defined as
\begin{equation}
    \x^{\star} = \argmin_{\x \in \mathcal{X}} f(\x).
\end{equation}
In a nutshell, BO algorithms build a probabilistic surrogate of $f$ based on its evaluations, such as a Gaussian process \citep{Jones1998} or a Bayesian neural network (BNN) \citep{snoek2015scalable, springenberg2016bayesian}, and sequentially select optimal new points to evaluate. The selection is guided by an acquisition function, which aims at balancing between exploration and exploitation.

Each domain-specific BO problem naturally has its domain experts with their own knowledge of the problem, i.e., often tacit knowledge about the shape of $f$ or where the global optimum might lie. Moreover, the cost of asking the expert can be significantly cheaper than the cost of obtaining the value of $f(\x)$ in many applications (e.g., when obtaining $f(\x)$ corresponds to running a simulation on a supercomputer or conducting a laboratory experiment). However, and surprisingly enough, leveraging that expert knowledge in order to speed up BO has only started to receive attention in the literature very recently. To the best of our knowledge, this problem has only been tackled from the perspective of incorporating expert prior knowledge about the location of the optimum into the BO acquisition function \citep{li2020incorporating,ramachandran2020incorporating,souza2021bayesian,hvarfner2021pi}. None of these works properly discuss how to obtain such a prior. The reason may be that eliciting knowledge from humans is notoriously challenging. Indeed, humans can be bad at evaluating absolute magnitudes, but on the other hand can be much better at comparing two instances \citep{shah2014better,millet1997effectiveness}. This has been utilized for learning of preferences through pairwise comparisons of items \citep{chu2005}, and has been expanded to an online learning setting \citep{brochu2008,gonzalez2017} to find the optimum of the preferences. Even though this approach has recently been shown to work in expert knowledge elicitation \citep{mikkola2020projective}, there is still a need for methods to elicit knowledge from the expert with the goal of performing BO for $f$. Secondly, transferring that knowledge into the BO task also represents a technical challenge.

In this paper, we propose an expert knowledge-augmented BO method, which solves the aforementioned issues. We formulate the problem as a multi-task learning (MTL) problem \citep{caruana1997multitask}, and propose to solve it with a Bayesian neural network-based architecture whose goal is to learn both $f$ and the expert knowledge. The key insight is to leverage statistical strength across the latent representations of the two tasks, which both are about the same ground truth $f$, even though both can be imperfect in different ways. We operate by first querying the expert, and then initialize the BO with that knowledge which leads to a speed-up for the BO. To make this architecture work, for the expert knowledge elicitation, we introduce a novel preference learning method based on Siamese neural networks. We call it a preferential Bayesian neural network (PBNN); it not only learns the instance preference relationship, but is also capable of capturing the latent function shape. As working with humans implies a limited number of queries, we use active learning to sequentially ask the most informative queries from the expert. Experiments demonstrate that PBNN leads to better performance than existing GP-based approaches with limited numbers of data acquisition steps. More importantly, we show that the standard BO optimization can be significantly sped-up when the elicited expert knowledge is transferred to the BO surrogate. For instance, if the simulated expert knowledge is $80\%$ accurate, the BO speedup varies between $1\times$ and $25\times$ across the experimented benchmark functions.

\section{Preferential Bayesian Neural Network}

In order to elicit the knowledge of the domain expert, we model it as a function, denoted by $g$, which represents the beliefs of the expert. We can interpret $g$ as a biased version of $f$. By querying pairwise comparisons from the expert, we build a probabilistic surrogate of $g$. Note that $g$ corresponds to the \textit{utility function} of the expert, which is a well-studied concept in economics \citep{rader1963existence}.

As motivated in the introduction, it is much easier for humans to compare two items than to give the absolute magnitude of one item (for instance it can be extremely difficult for a material scientist to compute a potential energy of a molecule, but comparing the stability between two molecules is easier). Hence, we assume that the expert cannot directly give the value $g(\x)$ for a certain $\x$. Instead, given a pair of covariates $[\x, \x'] \in \mathcal{X} \times \mathcal{X}$, we assume that the expert is able to return a preference label $\hat{y} \in \{0, 1 \}$, with value $\hat{y}_i = 1$ if $g(\x) \geq g(\x')$, and $\hat{y}_i = 0$ if $g(\x) < g(\x')$. We will sequentially collect a dataset $\mathcal{D}_g = \{ (\x_{i}, \x_{i}', \hat{y}_i)\}_{i=1}^{M}$, which is in turn used to build a probabilistic surrogate of $g$. Note that based on the ordered data, we will be able to learn about the shape of $g$, but not about the actual magnitude and scale, meaning that any monotonic transformation of $g$ is an equivalent solution.

\begin{figure}[t]
  \centering
  \centerline{\includegraphics[width=\textwidth]{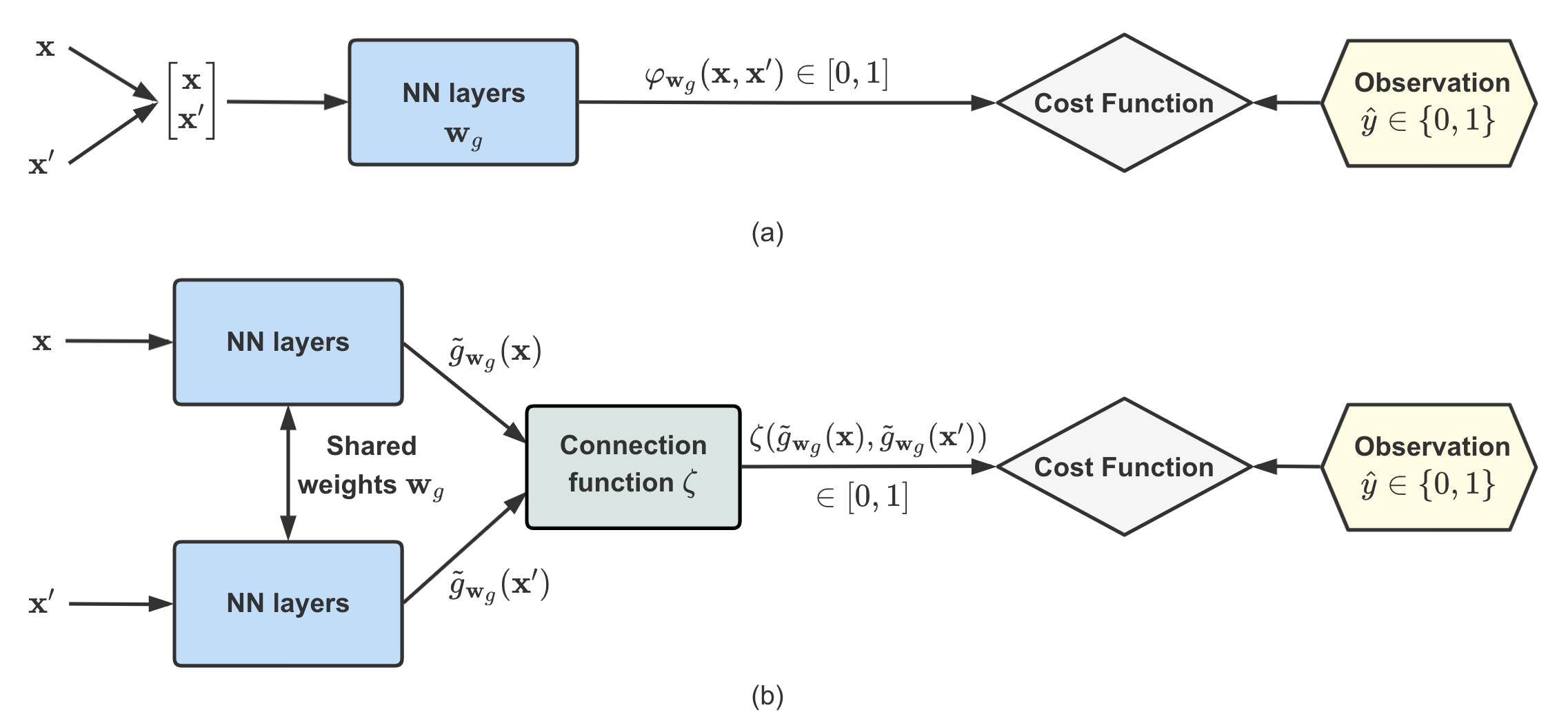}}
  \caption{(a) A simple neural network architecture to handle preference learning. The neural network outputs the probability of $\hat{y}$ to be 1 given $\x$ and $\x'$, but fails at capturing the shape of the utility function of the expert, $g$.
  (b) The proposed architecture, based on a Siamese neural network. It also solves the preference learning problem, but each sub-network outputs a real-valued latent representation that we interpret as $g(\x)$. The network is able to learn the shape of $g$, up to a monotonic transformation.
  }
  \label{fig:Siamese}
\end{figure}

We introduce a neural network-based architecture to handle preference learning and approach this problem as binary classification. More precisely, given two inputs $\x$ and $\x'$, we wish to output a value in $[0,1]$ that corresponds to the probability of $\hat{y}$ equaling 1. A natural solution to this problem would be to expand the input space to $\mathcal{X} \times \mathcal{X}$ by concatenating the covariates pair. Such an architecture is displayed in Figure~\ref{fig:Siamese}-a. However, by doing so, we would not learn anything about the function $g$. Instead, we propose to use an architecture based on Siamese networks (Figure~\ref{fig:Siamese}-b), coined PBNN (preferential Bayesian neural network), which is detailed in the next subsection.

\subsection{Preference Learning With Siamese Networks}

\subsubsection*{Network architecture and loss function}

A Siamese neural network consists of two parallel, identical sub-networks that share the same set of parameters. Each sub-network takes a distinct input, and the representations produced by each network are then compared using a connection function, which we denote by $\zeta$. They were introduced in the 90s for signature verification \citep{bromley1993signature}, and have since become very popular for, e.g., one-shot/few-shot learning \citep{koch2015siamese}, and object tracking \citep{bertinetto2016fully}. As our knowledge elicitation task amounts to a comparison between two values at a time, the Siamese network architecture naturally fits to our problem.

The proposed PBNN uses that architecture exactly. Let us denote by $\w_g$ the weights shared by the two sub-networks, and let us denote by $\tg_{\w_g}(\x)$ and $\tg_{\w_g}(\x')$ the representations produced by forwarding $\x$ and $\x'$. PBNN models the probability of $\hat{y}$ to be 1 given two inputs $\x$ and $\x'$ by comparing $\tg_{\w_g}(\x)$ and $\tg_{\w_g}(\x')$ with the connection function $\zeta$. Contrary to the ``concatenation'' baseline approach previously described (Figure~\ref{fig:Siamese}-a), the representations produced by the two sub-networks are real-valued, and we interpret them as the values of the true function $g$. These two values are further combined to provide a value in $[0,1]$, i.e., our connection function is naturally chosen as
\begin{equation}
    \zeta(\tg_{\w_g}(\x), \tg_{\w_g}(\x')) = \sigma(\tg_{\w_g}(\x) - \tg_{\w_g}(\x')),
\end{equation}
where $\sigma(x) = \dfrac{1}{1+e^{-x}}$ is the sigmoid function. We can then train the whole model by minimizing the negative log-likelihood, which is equivalent to using the binary cross-entropy loss. We write
\begin{align}
    \log p(\mathcal{D}_g|\w_g) & = \sum_{i=1}^{M} \log p(\hat{y}_i | [\x_i, \x_i'], \w_g) \\
    & = \sum_{i=1}^{M} \bigg( \hat{y}_i \log \big( \zeta(\tg_{\w_g}(\x), \tg_{\w_g}(\x')) \big)
    + (1-\hat{y_i}) \log \big( 1-\zeta(\tg_{\w_g}(\x), \tg_{\w_g}(\x')) \big) \bigg). \label{eq:pbnnloss}
\end{align}
To sum things up, the Siamese network-based architecture also solves the binary classification problem, but does so by learning an intermediate representation, which we identify as $g(\x)$. However, the current architecture only outputs a point estimate for $\tg_{\w_g}(\x)$, which is unsatisfying in our scenario where we wish to characterize uncertainties for active learning. To do so, we resort to Bayesian inference.

\subsubsection*{Bayesian inference}

To characterize the posterior distribution $p(\w_g|\mathcal{D}_g) \propto p(\mathcal{D}_g|\w_g) p(\w_g)$, the network weights $\mathbf{w}_g$ are equipped with a prior distribution $p(\w_g)$. The posterior is then in turn used to compute the predictive posterior distribution of $\tg_{\w_g}(\x)$. We resort to variational inference to characterize the posterior distribution. Variational inference aims at finding the closest approximation in terms of Kullback-Leibler divergence to $p(\w_g|\mathcal{D}_g)$, among a chosen family of distributions parameterized by $\bstheta_g$. Let us denote this approximation by $q(\w_g|\bstheta_g)$ (the so-called variational posterior). It can easily be shown that this amounts to minimizing the following expression w.r.t. $\bstheta_g$:
\begin{align} 
    \mathcal{L}_g(\bstheta_g) & = \mathrm{KL}[q(\w_g|\bstheta_g)||p(\w_g)] - \mathbb{E}_{q(\w_g|\bstheta_g)}[\log p(\mathcal{D}_g|\w_g)], \label{eq:elbo}
\end{align}
where the term $\log p(\mathcal{D}_g|\w_g)$ is given by Eq.~\eqref{eq:pbnnloss}. This expression is called the negative ELBO (evidence lower bound). The loss in Eq.~\eqref{eq:elbo} and its gradient are intractable, but we use Bayes by backprop (BBB) \citep{blundell2015weight} as our practical implementation. It provides Monte Carlo estimators of the loss and gradients, and ensures that back-propagation works. The minimization is then simply carried out by gradient descent. We refer the reader to the original paper for details. 

\begin{figure}[t]
  \centering
  \centerline{\includegraphics[width=\textwidth]{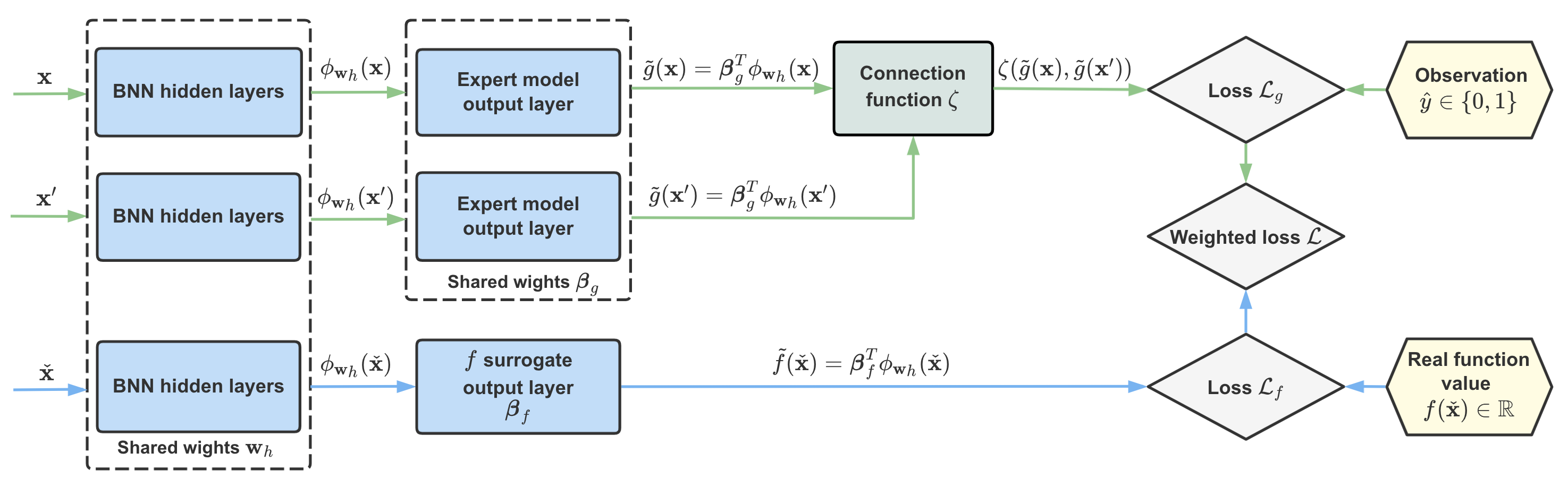}}
  \caption{Multi-task learning (MTL) architecture, for the tasks of building jointly probabilistic surrogates for both the expert's beliefs $g$ and the function to be optimized $f$. The green flow corresponds to surrogate of $g$, used in the knowledge elicitation part, and the blue flow to the surrogate of $f$, which is used for Bayesian optimization. The first layers (with parameters $\w_h$) are shared for the two surrogates, meaning that we aim at leveraging the similarity between the two functions by sharing some representations. They only differ in the output layer, parameterized by either $\bbeta_g$ or $\bbeta_f$. The two losses $\mathcal{L}_g$ and $\mathcal{L}_f$ are combined using a weighted scheme, which will give more and more importance to $\mathcal{L}_f$ as we get evaluations of $f$.}
  \label{fig:MTL}
\end{figure}

\subsection{Active Data Acquisition}  \label{sec:al}

Humans are not passive sources of information, and can only answer to a certain amount of queries before growing tired or impatient. In this limited budget setting, in order to maximize the use of the expert's time, we propose to resort to active learning to learn an accurate model in a sample-efficient way. Active learning methods sequentially select the most informative unlabeled instance (usually from a pool, which we denote by $\mathcal{D}_{pool}$), get the associated label, and retrain the model. Many different informativeness criteria have been proposed in the literature \citep{settles_active_2012}. Here, we propose to use the so-called BALD (Bayesian Active Learning by Disagreement, \citet{houlsby2011bayesian,gal2017deep}), a criterion justified from an information-theoretic perspective. BALD selects the point which maximizes the mutual information between its observation and model parameters. The optimal query $\x^{\star}$ is such that
\begin{align}
\x^{\star} & = \argmax_{x \in \mathcal{D}_{pool}}~\mathrm{I}(y; \w|\x, \mathcal{D}), \\
& = \Hh[y|\x, \mathcal{D}] - \mathbb{E}_{p(\w|\mathcal{D})}[\Hh[y|\x,\w]], \label{eq:bald}
\end{align}
where $\Hh$ denotes the differential entropy. Note that this is equivalent to maximizing the expected information gain on the model parameters, a well-known strategy in Bayesian experimental design \citep{lindley1956measure,mackay1992information,chaloner1995bayesian}.

Adapting BALD to our setting, we have
\begin{equation}
    [\x,\x']^{\star}=\argmax\limits_{[\x,\x'] \in \mathcal{D}_{pool}}~\mathrm{I} (\hat{y}; \w_g|[\x,\x'], \mathcal{D}_g).
\end{equation}
We approximate the mutual information as follows:
\begin{align} 
    & \mathrm{I}(\hat{y}, \w_g|[\x,\x'], \mathcal{D}_g) \notag \\
    & = \Hh[\hat{y}|[\x, \x'], \mathcal{D}_g] - \mathbb{E}_{p(\w|\mathcal{D}_g)}[\Hh[\hat{y}|[\x, \x'],\w_g]], \\
    & \simeq \Hh[\hat{y}|[\x, \x'], \mathcal{D}_g] - \mathbb{E}_{q(\w_g|\bstheta_g)}[\Hh[\hat{y}|[\x, \x'],\w_g]], \label{eq:pbald1} \\
    & \simeq h \left( \dfrac{1}{T} \sum_{t=1}^T \hat{p}_{\w_g^{(t)}}(\x, \x') \right) - \dfrac{1}{T} \sum_{t=1}^T h \left( \hat{p}_{\w_g^{(t)}}(\x, \x') \right), \label{eq:pbald2}
\end{align}
where we have used the notation
\begin{equation}
    \hat{p}_{\w_g}(\x, \x') = \zeta(\tg_{\w_g}(\x), \tg_{\w_g}(\x'))
\end{equation}
to denote the predicted probability that $\hat{y} = 1$ given the pair $[\x, \x']$ and parameters $\w_g$ (i.e., the output of the network with parameters $\w_g$), and where $h(p)= -p\log(p) - (1-p)\log(1-p)$ denotes the binary entropy function. The approximation in Eq.~\eqref{eq:pbald1} comes from swapping the true posterior distribution $p(\w_g|\mathcal{D}_g)$ with the variational posterior $q(\w_g|\bstheta_g)$, and the approximation in Eq.~\eqref{eq:pbald2} corresponds to Monte Carlo approximations given that the $\w_g^{(t)}$ are i.i.d. samples from $q(\w_g|\bstheta_g)$. We will refer to this criterion as PBALD.

\section{Expert Knowledge-Augmented Bayesian Optimization}\label{sec:mtl}

We now tackle the challenge of transferring what was learned in the previous step for the BO task. To that end, we propose to plug the previously described PBNN architecture into a wider multi-task learning (MTL) one. The MTL architecture aims at building probabilistic surrogates for both $f$ and $g$, by sharing the weights of the hidden layers, and having separate weights for the last layer. Indeed, we leverage the similarity between the functions $f$ and $g$ by sharing some of the latent representations produced by the network. The architecture is detailed in Section~\ref{sec-s-mtl}. As we are eliciting the knowledge of the expert in a first step, this will have the effect of providing the surrogate model for $f$ with a good initialization, which in turn will lead to accelerating the task of optimizing $f$. For this second step, we sequentially update this surrogate by collecting a dataset $\mathcal{D}_f = \{ \x_j, f(\x_j) \}_{j=1}^J$. Those points are selected using a BO acquisition function, as explained in Section~\ref{sec-s-acq}.

\subsection{Surrogate model for $f$} \label{sec:srrf}

The probabilistic surrogate we use for $f$ is a Bayesian neural network. Let us denote by $\w_f$ its weights, with prior distribution $p(\w_f)$. We further denote by $\tf_{\w_f}(\x)$ the output obtained by forwarding $\x$. Similarly, we resort to variational inference to characterize the posterior distribution $p(\w_f|\mathcal{D}_f)$, i.e., we aim at minimizing the following expression w.r.t. variational parameters $\bstheta_f$:
\begin{align}\label{eq:elbo_f}
    \mathcal{L}_f(\bstheta_f) =~& \mathrm{KL}[q(\w_f|\bstheta_f)||p(\w_f)] - \mathbb{E}_{q(\w_f|\bstheta_f)}[\log p(\mathcal{D}_f|\w_f)],
\end{align}
where $q(\w_f|\bstheta_f)$ denotes the variational posterior parameterized by $\bstheta_f$. Note that the log-likelihood term $p(\mathcal{D}_f|\w_f)$ is here Gaussian, which corresponds to the mean squared error.

A straightforward way of transferring what was learned in the first step would be to use the posterior distribution of the weights from the trained PBNN as the prior for $\w_f$. However, since PBNN does not learn the actual scale of $f$, using it to provide the prior distribution of the weights will not help at all, and may even lead to catastrophic forgetting \citep{french1999catastrophic}, where the shape information encoded in the posterior distribution of weights is erased during the training with $\mathcal{D}_f$.

To alleviate this problem, we consider a MTL architecture, with hard parameter sharing among the hidden layers for the surrogates of $f$ and $g$. In other words, we consider a joint model, whose shared parameters are going to be initialized through the trained PBNN. This is detailed next.

\subsection{Multi-task learning} \label{sec-s-mtl}

\subsubsection*{Parameter sharing}

We adopt hard parameter sharing among the weights of the hidden layers for the surrogates of $f$ and $g$. Let us split the weights $\w_g$ of PBNN into $\w_h$ and $\bbeta_g$, where $\w_h$ are the weights of all hidden layers and $\bbeta_g$ the weights of the output layer. That is, we can write $\tg(\x)=\bbeta_{g}^T\phi_{\w_h}(\x)$, where $\phi_{\w_h}(\x)$ represents the feature vector which is produced by forwarding $\x$ through all the hidden layers. The weights $\w_h$ are going to be shared for both surrogates, i.e., we can now write that the BNN surrogate of $f$ is parameterized by $\w_h$ and $\bbeta_{f}$, such that $\tf(\x) = \bbeta_{f}^T\phi_{\w_h}(\x)$ is the predicted outcome.

As such, the shared representation $\phi_{\w_h}(\x)$ will encode common features, such as the shape information that we wish to transfer to the surrogate of $f$. While expert knowledge may be biased, $\bbeta_{f}$ can be interpreted as a calibrator to lead the surrogate of $f$ to its actual scale and also rectify the potentially inaccurate information provided by the expert. By using the joint model, we will need fewer queries for Bayesian optimization, i.e., we save potentially expensive simulation costs.

\subsubsection*{Combining the losses}

The detailed architecture of the MTL system is presented in Figure~\ref{fig:MTL}. Now remains the question of combining the loss functions $\mathcal{L}_g$ (Eq.~\eqref{eq:elbo}) and $\mathcal{L}_f$ (Eq.~\eqref{eq:elbo_f}). In order to put more emphasis on the actual acquisitions of $f$ over time, we propose a weighted scheme with exponential decay for the $\mathcal{L}_g$. After the $j$-th BO acquisition, the loss $\mathcal{L}_j$ is such that
\begin{equation}\label{eq:combined}
    \mathcal{L}_j = \frac{\alpha^{j-1}}{\alpha^{j-1} + 1} \mathcal{L}_g + \frac{1}{\alpha^{j-1} + 1} \mathcal{L}_f,
\end{equation}
with $\alpha < 1$ a hyper-parameter to control the speed of the decay.

\begin{algorithm}[tb]
\caption{Expert knowledge-augmented BO}
\label{alg:algorithm}
    \textbf{Input}: 
    Active learning budget $M$,
    BO acquisition budget $J$ \\
    \textbf{Output}: Minimum of $f$
    \begin{algorithmic}[1] 
        \STATE \textbf{// Start Knowledge Elicitation}
        \STATE Initialize the expert model $\tg$ using PBNN with a random pair, $\mathcal{D}_g = \{(\x_0, \x_0', \hat{y}_0) \}$.
        \FOR {$i = 1 \text{ to } M$}
            \STATE $[\x_i, \x'_i] = \argmax\limits_{[\x,\x'] \in \mathcal{D}_{pool}} \mathrm{I} (\hat{y}; \w_g|[\x,\x'], \mathcal{D}_g)$ (Eq.~\eqref{eq:pbald2})
            \STATE Query the expert to obtain $\hat{y}_i$ associated to $[\x_i, \x'_i]$
            \STATE $\mathcal{D}_g \gets \mathcal{D}_g \cup (\x_i, \x'_i, \hat{y}_i)$
            \STATE Update variational parameters $\bstheta_g$ by minimizing Eq.~\eqref{eq:elbo}
        \ENDFOR
        
        \STATE \textbf{// Start Bayesian Optimization}
        \STATE $\mathcal{D}_f = \emptyset$
        \STATE $y_{best} = \infty$
        \FOR {$j = 1 \text{ to } J$}
            \STATE $\check{\x}_j = \argmax\limits_{\x \in \mathcal{X}}~\alpha_{\text{EI}}(\x)$
            \STATE Evaluate $f(\check{\x}_j)$
            \STATE $\mathcal{D}_{f} \gets \mathcal{D}_{f} \cup (\check{\x}_j, f(\check{\x}_j))$
            \STATE Update variational parameters $\bstheta_g$ and $\bstheta_f$ by minimizing the combined loss Eq.~\eqref{eq:combined}
            \STATE $y_{best} \gets \min (y_{best}, f(\argmin\limits_{\x \in \mathcal{X}}~\tilde{f}(\x)))$
        \ENDFOR
        
        \STATE \textbf{return} $y_{best}$
    \end{algorithmic}
\end{algorithm}

\subsection{Acquisition function} \label{sec-s-acq}

We adopt the expected improvement (EI) as the acquisition function \cite{Jones1998}. Given $\mu_{\x}$ the predictive mean of BO surrogate model and $s^2_{\x}$ the predictive variance, the EI at point $\x$ can be defined as:
\begin{eqnarray} \label{eq:ei}
    \alpha_{\text{EI}}(\x) = s_{\x}[\gamma(\x)\Phi(\gamma(\x))+ \psi(\gamma(\x))],
\end{eqnarray}
where $\gamma(\x) = (y_{best} - \mu_{\x})/s_{\x}, y_{best}$ is the current lowest value of the objective function, and the $\Phi(\cdot)$, $\psi(\cdot)$ are the cumulative distribution function and probability density function of a standard normal random variable. Since Eq.~\eqref{eq:ei} is intractable for a BNN as there is no analytical form of the output distribution, we use Monte Carlo sampling to obtain the approximate EI \citep{kim2021scalable}:
\begin{eqnarray} \label{eq:mcei}
    \alpha_{\text{EI}}(\x) \approx \dfrac{1}{T}\sum_{t=1}^{T} \max\left(y_{best}-\tilde{f}^{(t)}(\x), 0\right),
\end{eqnarray}
where the $\tilde{f}^{(t)}(\x)$ are i.i.d. predictive samples at $\x$.

Our two-step, expert knowledge-augmented BO procedure (knowledge elicitation first, and BO second) is summed up in Algorithm~\ref{alg:algorithm}.

\section{Experiments}\label{sec:experiment}

\subsection{Performance of the PBNN architecture}

\subsubsection{Toy example: Capturing the shape}

\begin{figure}[t]
    \centering
    \includegraphics[width=0.7\linewidth]{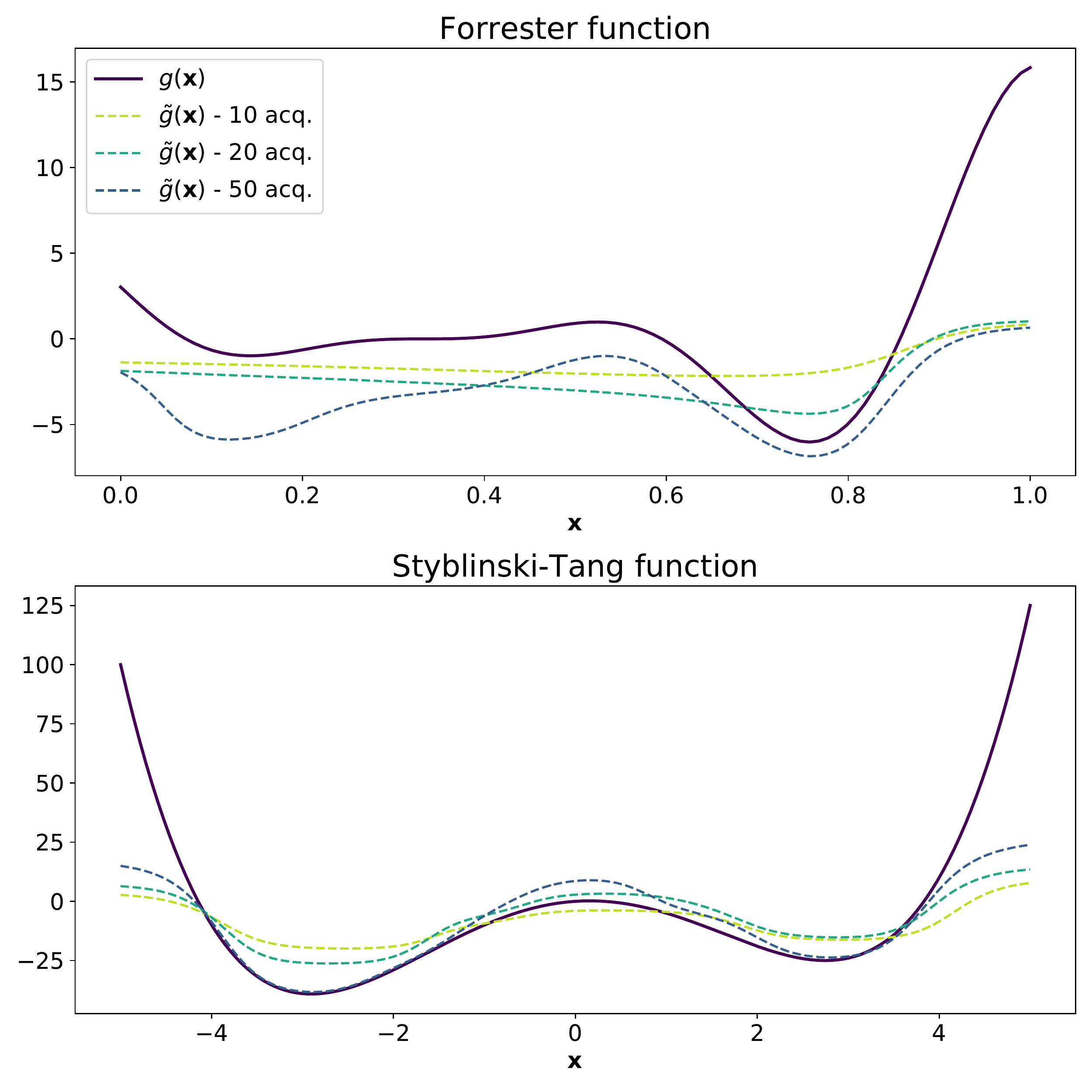}
    \caption{Toy example illustrating how the proposed PBNN architecture can learn the shape of a function using pairwise comparisons. Experiments carried out on the Forrester (top) and Styblinski-Tang (bottom) functions. The thickest, dark blue line represents the true function $g$, while the dotted lines represent the predicted functions $\tg$ learned by PBNN using 10, 20 and 50 pairwise comparisons.}
    \label{fig:shape}
\end{figure}

We first present a toy example using 1-dimensional benchmark functions to illustrate how the proposed PBNN architecture can learn the shape of the function $g$ through different number of pairwise comparisons. The model is trained by sequentially selecting 10, 20 and 50 pairwise comparisons using the PBALD criterion and getting the associated preference labels. We assume noiseless feedback in this experiment for illustration purposes. Figure~\ref{fig:shape} displays the comparison between the real function values $g$ and elicited expert model predictions $\tg$, with the Forrester and Styblinski-Tang functions. From the figure, it can been seen that with 50 pairwise comparisons, the expert model $\tg$ can capture the ordinal information, i.e., $g$ up to a monotonic constant, as previously explained.

\subsubsection{Elicitation performance}

We compare the performance of PBNN with the classical GP-based preference learning model by \cite{chu2005} on three different datasets. We artificially transform three regression datasets into preference datasets by creating preference labels between all possible pairs of covariates using the target values.

For the GP-based model, we use the squared exponential kernel with the hyperparameters optimized by maximum marginal likelihood. The PBNN architecture consists of two fully connected layers with width $100$ and $10$, where each hidden layer is followed by a \texttt{tanh} activation function. Optimization is carried out using ADAM \citep{kingma2014adam} with learning rate 0.001. We use a Gaussian variational posterior with $\mathcal{N}(0, 0.1)$ as priors for the weights. 

The models are initialized using one random pair, and then a total of $N_{AL}$ pairs are sequentially queried by maximizing the BALD criterion \citep{houlsby2011bayesian} for the GP-based model, and the PBALD criterion for PBNN, respectively. After the active learning phase, the accuracy of the model is assessed by computing a binary accuracy score on a hold-out test set consisting of 1000 pairs. The results on the three datasets, averaged over 20 replications, are presented in Table~\ref{tab:pref} for $N_{AL} = 50$ and $N_{AL} = 100$. In all scenarios, PBNN achieved better accuracy results w.r.t. the GP-based model.

Lastly, we propose to compare the runtimes of the two methods. The inference of PBNN is typically run GPU-based architectures, however, for a fair comparison, we compare 20 runs of each method on the three datasets using $N_{AL} = 50$ using the same CPU architecture\footnote{2x20 core Xeon Gold 6248 2.50GHz, 192GB RAM.}. The results are reported in Table \ref{tab:runtimes}. On all three datasets, the proposed PBNN is roughly 20 times faster than the GP-based baseline.

\begin{table}[t]
    \scriptsize
    \centering
    \caption{Accuracy of preference prediction after $N_{AL}$ acquisitions is significantly better than the earlier GP-based method on three datasets. The accuracy score corresponds to the proportion of correct binary preference prediction on a hold-out test set comprising 1000 pairs. The mean and standard deviation of this score are reported over 20 replications.}
    \label{tab:pref}
        \begin{tabular}{l|c|cc}
        \hline
                                               &             &\multicolumn{2}{|c}{Accuracy (\%)}   \\
        \hline
        DATASET                                & $N_{AL}$    & GP & PBNN \\
        \hline
        \multirow{2}{*}{Machine CPU (6D)}      & 50      & $67.29 \pm 2.91$  & $\mathbf{81.07} \pm                                                                               \mathbf{3.74}$     \\
                                               & 100     & $69.45 \pm 1.82$  & $\mathbf{84.38} \pm                                        \mathbf{1.52}$     \\
        \hline
        \multirow{2}{*}{Boston Housing (13D)}  & 50      & $67.41\pm2.35$    & $\mathbf{83.40} \pm                                                                               \mathbf{2.14}$     \\
                                               & 100     & $69.33\pm2.50$    & $\mathbf{85.89} \pm                                        \mathbf{2.52}$     \\
        \hline
        \multirow{2}{*}{Pyrimidine (27D)}      & 50      & $70.99 \pm 2.35$  &  $\mathbf{79.63} \pm                                                                              \mathbf{2.14}$    \\
                                               & 100     & $79.29\pm 2.50$   &  $\mathbf{86.59} \pm                                       \mathbf{2.52}$    \\
        \hline
        \end{tabular}
\end{table}

\begin{table}[t]
    \centering
    \caption{Average runtimes (in seconds) after $N_{AL} = 50$ acquisitions on three datasets. The runtime of the proposed PBNN is roughly 20 times faster than the GP-based method. The comparison was carried out with 20 runs on the same CPU architecture. The standard deviation is also reported.}
    \label{tab:runtimes}
    \begin{tabular}{|ll|cc}
    \hline
    \multicolumn{1}{l|}{}    &              & \multicolumn{2}{c}{Runtimes (sec.)}                  \\ \hline
    \multicolumn{1}{l|}{DATASET} & $N_{AL}$ & GP                       & PBNN                    \\ \hline
    \multicolumn{1}{l|}{Machine CPU (6D)} & 50       & \multicolumn{1}{c}{899 $\pm$ 68} & \multicolumn{1}{c}{$\mathbf{40 \pm 4}$} \\ \hline
    \multicolumn{1}{l|}{Boston Housing (13D)}  & 50       & \multicolumn{1}{c}{981 $\pm$ 72} & \multicolumn{1}{c}{$\mathbf{50 \pm 7}$} \\ \hline
    \multicolumn{1}{l|}{Pyrimidine (27D)} & 50 & \multicolumn{1}{c}{1003 $\pm$ 64} & \multicolumn{1}{c}{$\mathbf{57 \pm 9 }$} \\ \hline
    \end{tabular}
\end{table}

\subsection{Comparison of the optimization performance on benchmark functions}

\subsubsection{Experiment with simulated experts} \label{sec:exp_sim}

\begin{figure}[t]
  \centering
  \includegraphics[width=0.7\linewidth]{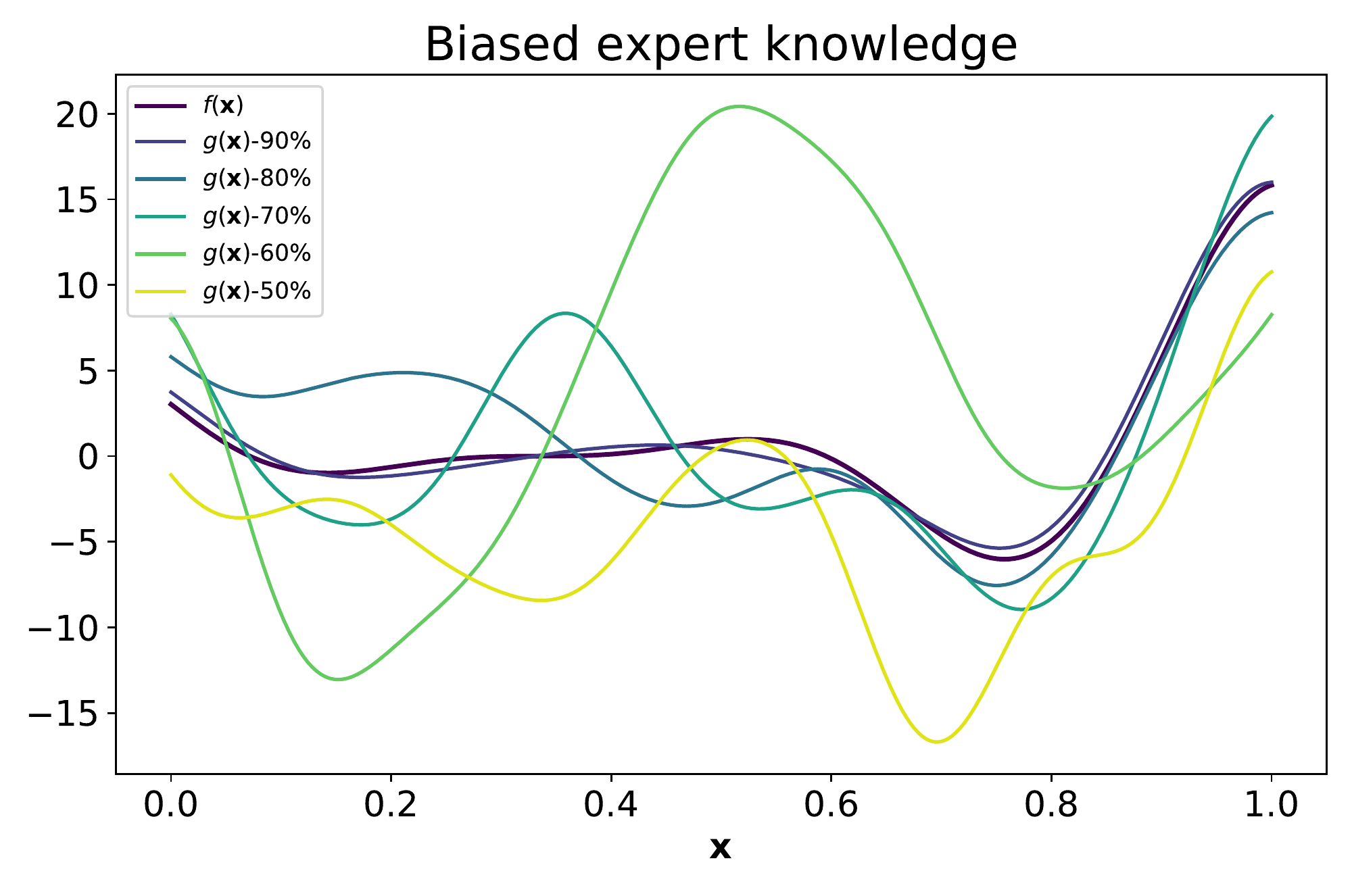}
  \caption{Illustration of the simulated expert's beliefs using the Forrester function. The true function is the thickest, dark blue curve, and the other curves correspond to that function perturbed with a random GP draw with various variances. The variances are chosen such that the accuracy of the expert ranges from 50\% to 90 \%.}\label{fig:expert}
\end{figure}

We first study the performance of the proposed knowledge-augmented Bayesian optimization scheme in a simulated setting where we can control the bias of an expert. More precisely, we compare how well the scheme performs w.r.t. to standard Bayesian optimization on several benchmark functions from the literature.

We assume an expert with potentially biased beliefs of the true function $f$, with the bias expressed as a perturbation function $\delta$:
\begin{align}
    g(x) = f(x) + \delta(x),
\end{align}
where $\delta$ is a zero-mean Gaussian process draw with kernel $\sigma_{\delta}^2 k(x,x')$, which encodes the form of the expert's bias. Note that this does not reduce generality, assuming a general enough perturbation family. In the experiments reported below, we study the effect of expert’s bias on the performance by choosing the SE kernel with lengthscale $\ell=0.1$ and varying $\sigma_{\delta}^2$ so that we obtain five levels of expert knowledge accuracy from 50\% up to 90\%. Figure~\ref{fig:expert} illustrates the comparison between the simulated experts with the ground truth objective function with the Forrester function, for the 5 considered accuracy levels.

\begin{figure}[t]
  \centering
  \centerline{\includegraphics[width=\textwidth]{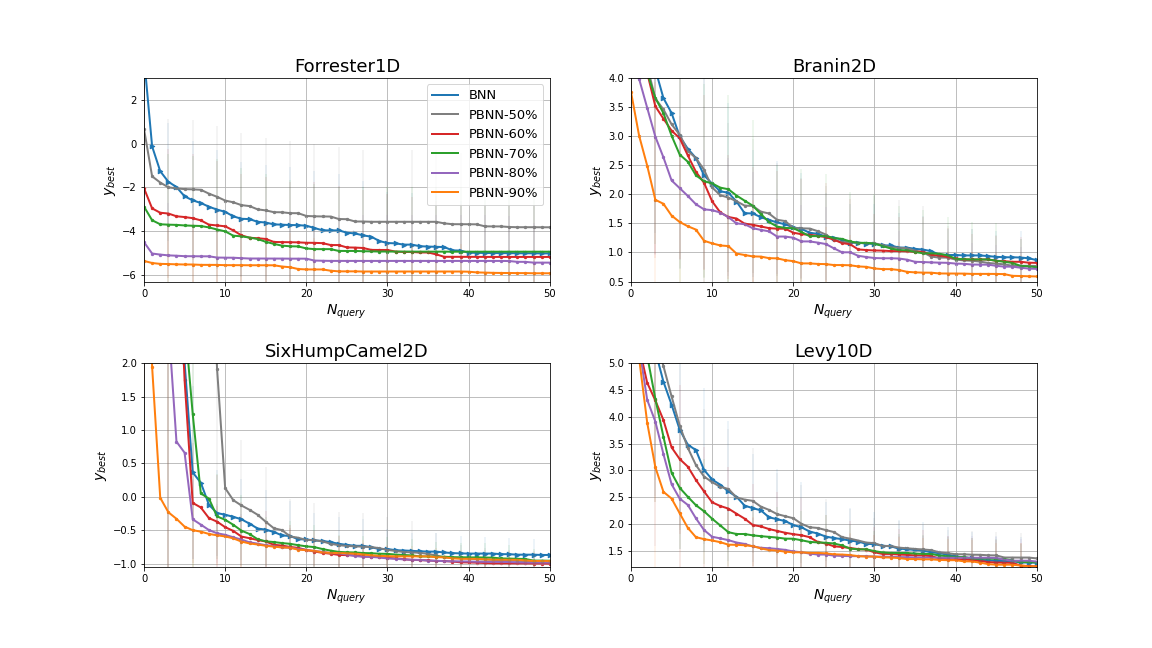}}
  \caption{Comparison of the optimization performance of the expert knowledge-augmented BO using 4 benchmark functions w.r.t. standard BO. We simulate 5 experts with different levels of knowledge (denoted \texttt{PBNN-xx\%}), where the percentage stands for the accuracy of the expert's preferential feedback, i.e., $50\%$ means that the expert is simulated with fully biased knowledge. The knowledge of the simulated expert is elicited using $M=100$ pairwise comparisons. The standard BO scheme, i.e., without expert knowledge, is denoted \texttt{BNN}. The results are averaged over 50 simulations.}\label{fig:bo}
\end{figure}

For the MTL structure, the shared hidden layers have width [100, 30, 15]. Standard BO is run using the BNN surrogate described in Section~\ref{sec:srrf}, in other words, it is the exact same architecture as the ``BO branch'' of the MTL, for fair comparison. Experiments were run with $M = 100, J = 50$ and $\alpha = 0.95$. The results, detailed in the next paragraph, are averaged over 50 replications of the experiment. The full description of experimental settings is provided in Appendix~\ref{appB}.

Figure~\ref{fig:bo} shows the results on four benchmark functions\footnote{https://www.sfu.ca/~ssurjano/optimization.html}: ``Forrester1D'', ``Six-hump-camel2D'', ``Branin2D'' and ``Levy10D''. The results are evaluated by $y_{best}$, which is the current minimal value of the true objective function predicted by the surrogate of $f$. We can see that the more accurate the simulated expert is, the more pronounced the acceleration effect. If the expert is reliable enough, we can speed up BO significantly. When the expert does not have any knowledge, i.e. 50\% preference accuracy, this actually leads to performance deterioration w.r.t. standard BO, which meets our expectation. For the all expert accuracy levels $\geq 60\%$, the final round BO performance ($N_{query} = 50$) is at least as competitive as the standard BO. However, the gain is much more striking in the early stages of the BO ($N_{query} \ll 50$). This phenomenon may be due to the challenge of making use of inaccurate expert knowledge as more accurate ground-truth data becomes increasingly available. We also provide additional experiment results with different parameter settings in Appendix~\ref{appC}.

\subsubsection{Experiment with human experts}

We further study the performance of the proposed method in a real-world setting with actual human experts. To that end, we conduct a simple user experiment involving memory abilities. Similarly to the previous experiment, the goal is to optimize BO benchmark functions. To induce controlled knowledge about those functions, we let users memorize the shape of the objective function by displaying 2D-plots for a short time. This provides useful but biased preference information for optimization. For the experiment setup, we restrict the objective functions to 2D benchmark functions as we cannot properly display 3D (or more) functions to users. Based on their memory of the function, the user must then answer a series of questions asked in preferential form, in asked in the format ``\textit{At which point do you think the value of the function is larger?}''. The questions are determined by the PBALD criterion, detailed in Section~\ref{sec:al}. Finally, BO augmented with expert knowledge is then run with the proposed methodology. We compare this approach with standard BO. The intuition behind this experiment is that users cannot memorize the function in all its complexity, but still can grasp an understanding its overall shape, which could speed up the optimization.

We choose three commonly used 2D benchmark functions: "Six-hump-camel2D", "Three-hump-camel2D" and "Branin2D". This choice is motivated by the fact that these particular functions have several local minima, but are still smooth enough so that users can remember their general shape in a short time. The plots are displayed for 2 minutes; we believe this is enough time to remember the general shape of the function, but not learn perfect information, thus mimicking expert knowledge on complex problems. The number of preferential questions is set to 25, a number not too small to effectively build the expert model nor too large to bore the users. The coordinates of the points selected by each question, as well as their locations in the coordinate system used for the visualization of the function, are provided to the user. Other settings remain the same as the experiments of Section~\ref{sec:exp_sim}. We also provide a brief instruction manual for the users (see Appendix~\ref{appA}). The experiment is carried out following an existing code of conduct about user studies. The users are recruited from a student population that have no previous knowledge of the test functions.  

\begin{table}[t]
\caption{User accuracy for the three BO benchmark functions, i.e., the percentage of questions each user answered correctly (out of 25).}
\centering
\label{tab:expert_acc}
\small{
\begin{tabular}{l|lllllllll}
\hline & \multicolumn{9}{c}{Accuracy} \\ \hline
Function (2D)  & \multicolumn{1}{l|}{User1} & \multicolumn{1}{l|}{User2} & \multicolumn{1}{l|}{User3} & \multicolumn{1}{l|}{User4} & \multicolumn{1}{l|}{User5} & \multicolumn{1}{l|}{User6} & \multicolumn{1}{l|}{User7} & \multicolumn{1}{l||}{User8} & \textbf{Avg.} \\ \hline
3H-camel & \multicolumn{1}{l|}{64\%}     & \multicolumn{1}{l|}{84\%}     & \multicolumn{1}{l|}{72\%}     & \multicolumn{1}{l|}{80\%}     & \multicolumn{1}{l|}{52\%}     & \multicolumn{1}{l|}{80\%}     & \multicolumn{1}{l|}{72\%}     & \multicolumn{1}{l||}{68\%}     & \textbf{71.5\%}         \\ \hline
6H-camel   & \multicolumn{1}{l|}{52\%}     & \multicolumn{1}{l|}{68\%}     & \multicolumn{1}{l|}{76\%}     & \multicolumn{1}{l|}{88\%}     & \multicolumn{1}{l|}{64\%}    & \multicolumn{1}{l|}{84\%}     & \multicolumn{1}{l|}{72\%}     & \multicolumn{1}{l||}{56\%}     & \textbf{70\%}           \\ \hline
Branin         & \multicolumn{1}{l|}{80\%}     & \multicolumn{1}{l|}{72\%}     & \multicolumn{1}{l|}{72\%}     & \multicolumn{1}{l|}{80\%}     & \multicolumn{1}{l|}{68\%}     & \multicolumn{1}{l|}{84\%}     & \multicolumn{1}{l|}{80\%}     & \multicolumn{1}{l||}{76\%}     & \textbf{76.5\%}         \\ \hline
\end{tabular}
}
\end{table}

\begin{figure}[h!]
  \centering
  \centerline{\includegraphics[width=0.7\textwidth]{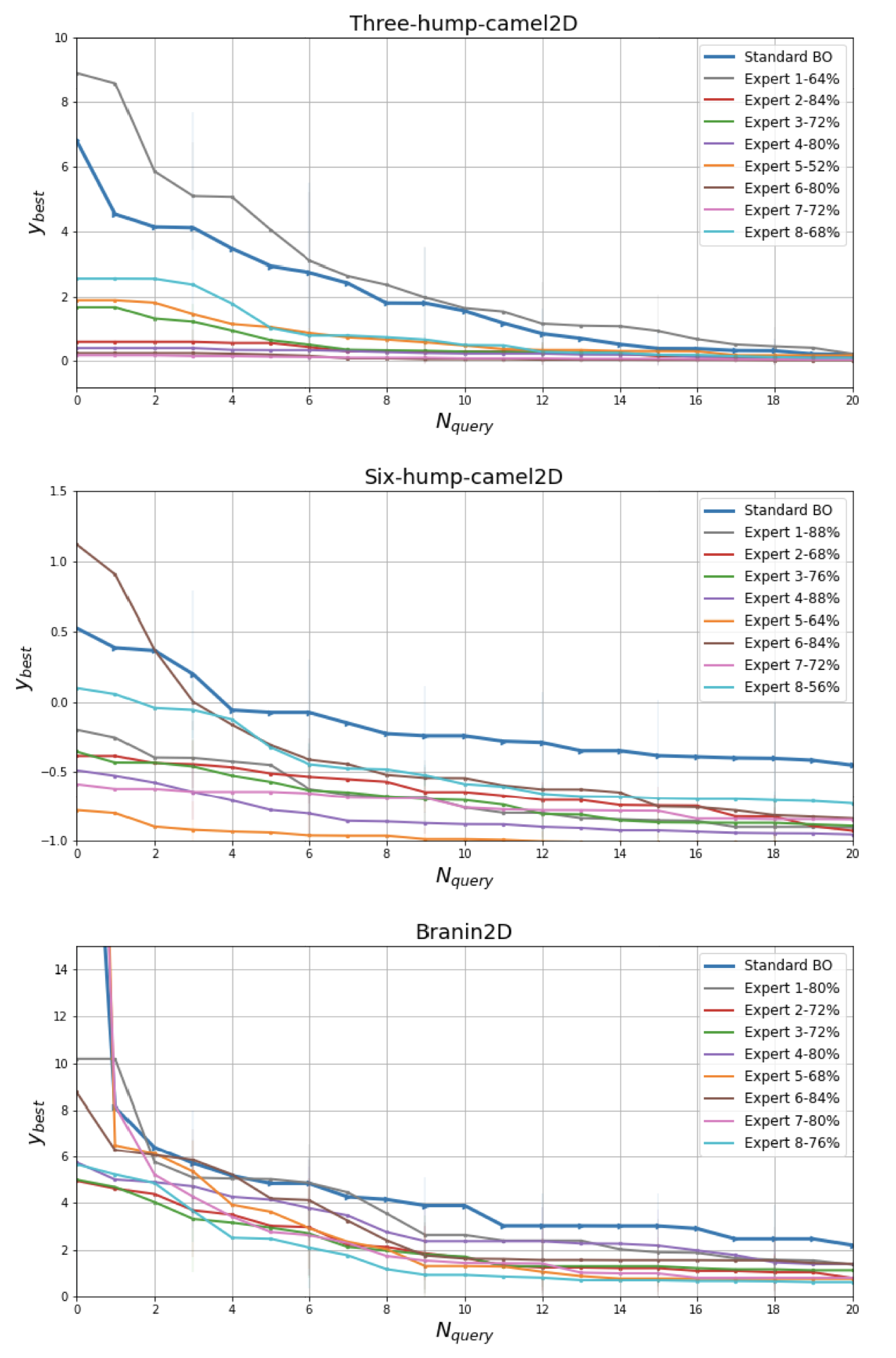}}
  \caption{Comparison of the optimization performance of the expert knowledge-augmented BO with real users on 3 2D benchmark functions w.r.t. standard BO. We collect the data from 8 users, where the percentage stands for the accuracy of the expert’s preferential feedback. The knowledge of the simulated expert is elicited using M = 25 pairwise comparisons. The results are averaged over 10 simulations.}
  \label{fig:real_experiment}
\end{figure}

Table~\ref{tab:expert_acc} reports the accuracy of eight different users on the three selected objective functions. All the accuracy rates are greater than 50\%, and the average accuracy is 70\% or higher for each function, which is useful information. Figure~\ref{fig:real_experiment} shows the comparison between standard BO and our method in terms of optimization performance. Each simulation builds the expert model using PBNN with the same dataset obtained from each user, but with different network initialization. We run 10 simulations to account for the randomness. As can be seen on those plots, the help of experts leads to prominent acceleration compared with standard BO, which again proves the effectiveness of our expert knowledge-augmented BO method.

\section{Conclusion}

In this paper, we tackled the incorporation of human expert knowledge into BO with the goal of speeding up the optimization. Our procedure breaks down into two steps. The first is to elicit the expert beliefs by querying them with pairwise comparisons. By doing so, we obtain the approximate shape of the objective function. The second step is to share the expert knowledge with the BO, to provide auxiliary information about the potential location of the optimum.

More precisely, we proposed PBNN, a novel preference learning architecture based on Siamese networks to efficiently elicit the expert knowledge. By sequentially querying the preferences between two objects with active learning, the proposed PBNN is more powerful in capturing the latent preference relationships compared with the former GP-based model on different datasets. To conduct the knowledge transfer, we design a well-aligned multi-task learning structure with a knowledge sharing scheme to combine our expert model with BO surrogate. Experiments on different benchmark functions show that when the expert is trustworthy, we can gain significant benefit from the elicited knowledge and markedly speed up the optimization.

A limitation of this paper is that we conducted the knowledge elicitation blindly w.r.t. the task of optimizing $f$. We will aim at proposing task-oriented active learning criterion in future work. Another exciting direction for future work is to directly consider the expert as another source of information, and therefore not resort to the two-step approach considered here. This would bring this work closer to the multi-fidelity BO line of research \citep{kandasamy2016gaussian,takeno2020multi,li2020incorporating}. In that scenario, as humans are not passive sources of information, but rather active planners, we would need to build models able to anticipate behaviours such as steering \citep{colella2020human}.

\clearpage

\bibliography{bibliography}
\bibliographystyle{apalike}

\clearpage

\appendix

\section{User Manual}\label{appA}

\subsection{Introduction}
Welcome to the test. During this experiment, you need to try your best to remember the shapes of three different 2-D functions in limited time. After that you need to answer 25 simple questions, by telling which point do you think is larger between a pair of points.

In this test, we rely on the existing code of conduct for conducting user studies in our field. The experimental data is only used for this paper, and we will not disclose any of your private information.

\subsection{Experimental details}
\begin{figure}[H]
  \centering
  \centerline{\includegraphics[width=0.7\textwidth]{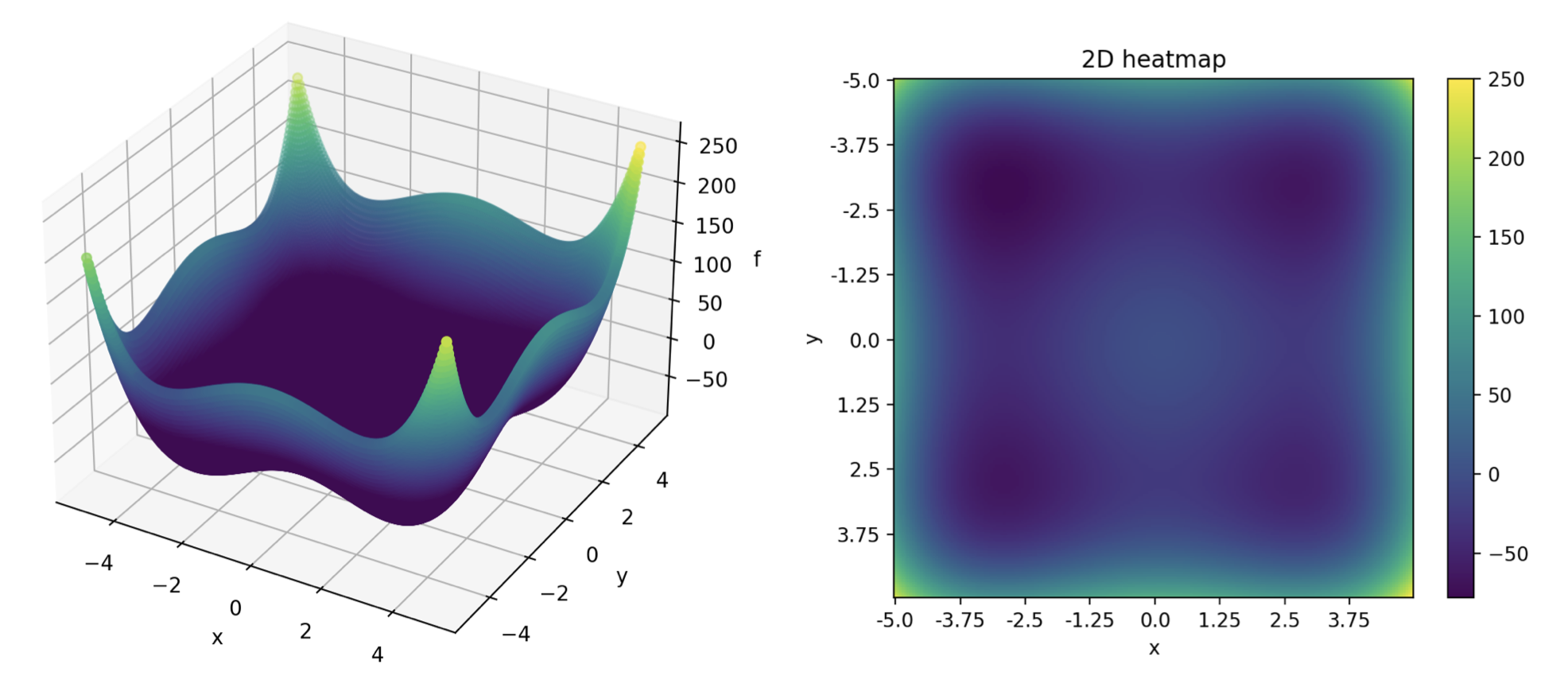}}
  \caption{The left plot is the 3-D view of the objective function, and the right one is the 2-D heat map. These plots are only for demonstration, the real objective functions in the experiment will not be shown here.}
  \label{fig:plot_demo}
\end{figure}

Three experiments will be conducted in random order. When each experiment starts, you will be shown a 3-D plot and a 2-D heat map of the function (demo plots are shown in Figure~\ref{fig:plot_demo}), and you can drag the 3-D plot to have a better visualization. You will have 2 minutes to remember the plots, once the time is up, you will no longer be able to view these plots.

\begin{figure}[H]
  \centering
  \centerline{\includegraphics[width=0.45\textwidth]{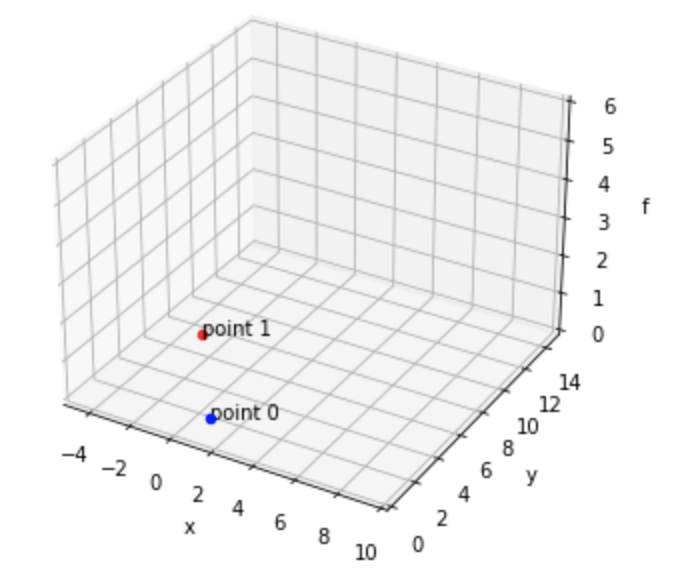}}
  \caption{The plot of two points in the question stage.}
  \label{fig:question_plot}
\end{figure}

After that, you will be asked 25 questions. Each question is asked in the format "At which point do you think the value of the function is larger?" And there will be no time limit for you to answer these questions. We will provide the coordinates of the two points to you and also plot their locations in the coordinate system used in the visualization of the function. The demo plot is shown in Figure~\ref{fig:question_plot}.

\begin{figure}[H]
  \centering
  \centerline{\includegraphics[width=0.45\textwidth]{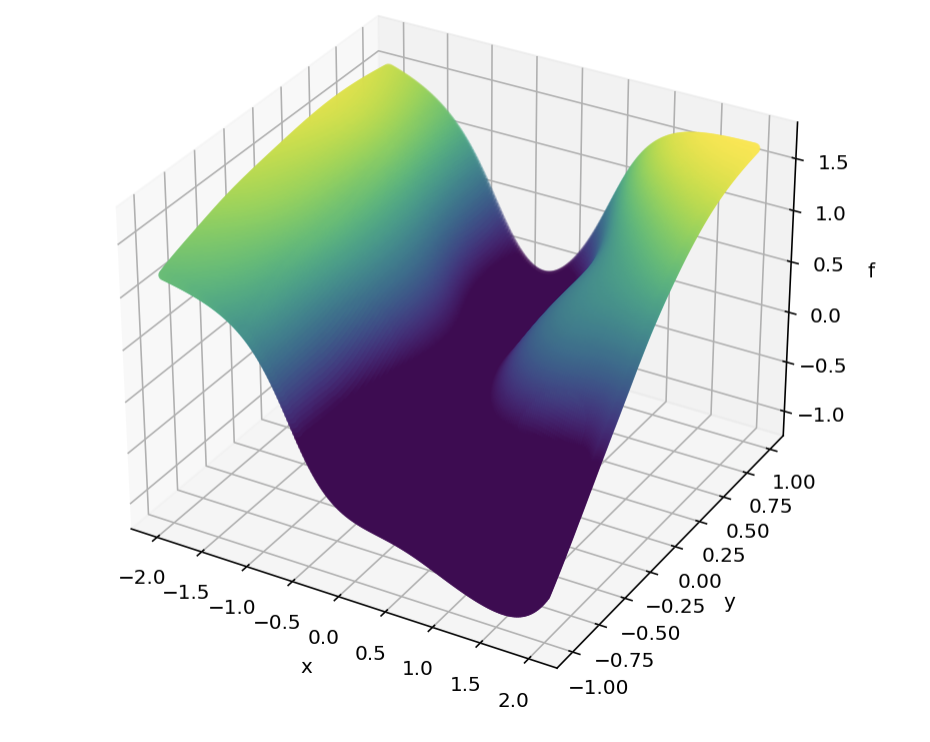}}
  \caption{An example of the user model}
  \label{fig:user_model}
\end{figure}

After answering all the questions, you will directly jump to the next experiment. Upon finish the three experiments, the system will calculate the accuracy of your performance, and you can also view your own user model in 3D plot (see Figure~\ref{fig:user_model} for reference).

\section{Experimental settings} \label{appB}
\subsection{Elicitation experiment}

\subsubsection*{Datasets}
\begin{itemize}
	\item Machine CPU: A computer hardware dataset. The dimension is 6 and has 209 instances
	\item Boston housing: This dataset contains information concerning housing in the area of Boston Mass. The dimension is 13 and has 506 cases
	\item Pyrimidine: A pyrimidine QSAR dataset. The dimension of this dataset is 27 and has 74 instances
\end{itemize}

The initial training set contains one random pair. The query pool size for active learning is 2000 pairs, and the test set used for evaluating accuracy consists of 1000 pairs. The dataset is shuffled in each epoch. 

\subsubsection*{Hyper-parameters}
\begin{itemize}
    \item Number of active acquisitions in elicitation stage: 50, 100
    \item Monte Carlo sampling budget in BALD: 100
    \item Number of simulations: 20
\end{itemize}

\subsubsection*{Neural network configurations}
\begin{itemize}
    \item Framework: PyTorch, torchbnn
	\item Optimizer: ADAM with learning rate = 0.001
	\item Scheduler: CosineAnnealingLR with $T_{max} = 20$ and $eta_{min}=0.0001$
	\item Batch size: 2
	\item Number of epochs: 20
	\item Bayesian linear layer: 2 layers with weight prior $\mathcal{N}(0, 0.1)$, width [100, 10]
	\item Activation function: Tanh
\end{itemize}

\subsection{BO with simulated experts} \label{sec:bo_setting}
\subsubsection*{Benchmark functions}
\begin{itemize}
	\item Forrester1D: A simple one-dimensional test function, with one global minimum, one local minimum and a zero-gradient inflection point. This function is evaluated on $x \in [0, 1]$. The form of this function is:
    	\begin{equation}
    	    f(x) = (6x-2)^2 \sin(12x-4).
    	\end{equation}
	\item Branin2D: A 2D function with three global minima. We take $a = 1, b = \dfrac{5.1}{4 \pi^2}, c = \dfrac{5}{\pi}, r = 6, s = 10  \ \mathrm{and} \ t = \dfrac{1}{8\pi}$. This function is evaluated on the square $x_1 \in [-5, 10], x_2 \in [0, 15]$. The function form is:
    	\begin{equation}
    	    f(x) = a(x_2-bx_1^2+cx_1-r)^2+s(1-t)\cos(x_1)+s.
    	\end{equation}
	\item Six-hump-camel2D: A 2D function with six local minima, two of which are global. This function is evaluated on the square $x_1 \in [-3, 3], x_2 \in [-2, 2]$. The function form is:
    	\begin{equation}
    	    f(x) = (4-2.1x_1^2+\dfrac{x_1^4}{3})x_1^2+x_1 x_2+(-4+4x_2^2)x_2^2.
    	\end{equation}
	\item Levy10D: A 10D function evaluated on the hypercube $x_i \in [-2, 2]$, for all $i= 1, ..., d$. The function form is:
    	\begin{align}
    	    f(x) =~& \sin^2(\pi w_1) + \sum_{i=1}^{d-1}(w_i -1)^2[1+10\sin^2(\pi w_i + 1)] \\ \nonumber
    	    & + (w_d-1)^2[1+\sin^2(2\pi w_d)], 
    	\end{align}
    	 where $w_i = 1 + \dfrac{x_i -1}{4}$, for all $i = 1,...,d$.
\end{itemize}

The number if initial training pairs for elicitation is 1. The query pool size for active learning is 2000. 

\subsubsection*{Hyper-parameters}
\begin{itemize}
    \item Number of active acquisitions in elicitation stage: 100
    \item Number of BO acquisition: 50
    \item Monte Carlo sampling budget in BALD: 100
    \item Monte Carlo sampling budget in EI: 30
    \item $\alpha$ in MTL shared weight: 0.95 
    \item Number of simulations: 50
\end{itemize}

\subsubsection*{Neural network configurations}
\begin{itemize}
    \item Framework: PyTorch, torchbnn
	\item Optimizer: ADAM with lr = 0.001 in elicitation stage, lr = 0.01 in BO stage
	\item Scheduler: CosineAnnealingLR with $T_{max} = 20$ and $eta_{min}=0.0001$ in elicitation stage, no scheduler in BO.
	\item Batch size: 10 for preference data, 5 for regression data
	\item Number of epochs: 100 in elicitation stage, 200 in BO stage
	\item Bayesian linear layer: 3 shared hidden layers with weight prior $\mathcal{N}(0, 0.1)$, width [100, 30, 15]
	\item Activation function: Tanh
\end{itemize}

\subsection{BO with actual human experts}
\subsubsection*{Benchmark functions}
\begin{itemize}
	\item Three-hump-camel2D: This function has three local minima and is evaluated on the square $x_1 \in [-2, 2], x_2 \in [-2, 2]$. The form of this function is:
    	\begin{equation}
    	    f(x) = 2x_1^2-1.05x_1^4+\dfrac{x_1^6}{6}+x_1x_2+x_2^2.
    	\end{equation}
	
	\item Six-hump-camel2D: A 2D function with six local minima, two of which are global. This function is evaluated on the square $x_1 \in [-2, 2], x_2 \in [-1, 1]$. The function form is:
    	\begin{equation}
    	    f(x) = (4-2.1x_1^2+\dfrac{x_1^4}{3})x_1^2+x_1 x_2+(-4+4x_2^2)x_2^2.
    	\end{equation}
    	
	\item Branin2D: A 2D function with three global minima. We take $a = 1, b = \dfrac{5.1}{4 \pi^2}, c = \dfrac{5}{\pi}, r = 6, s = 10  \ \mathrm{and} \ t = \dfrac{1}{8\pi}$. This function is evaluated on the square $x_1 \in [-5, 10], x_2 \in [0, 15]$. The function form is:
    	\begin{equation}
    	    f(x) = a(x_2-bx_1^2+cx_1-r)^2+s(1-t)\cos(x_1)+s.
    	\end{equation}
\end{itemize}

The number if initial training pairs for elicitation is 1. The query pool size for active learning is 2000. 

\subsubsection*{Hyper-parameters}
\begin{itemize}
    \item Number of active acquisitions in elicitation stage: 25
    \item Number of BO acquisition: 20
    \item Monte Carlo sampling budget in BALD: 100
    \item Monte Carlo sampling budget in EI: 30
    \item $\alpha$ in MTL shared weight: 0.95 
    \item Number of simulations: 10
\end{itemize}

\subsubsection*{Neural network configurations}
\begin{itemize}
    \item Framework: PyTorch, torchbnn
	\item Optimizer: ADAM with lr = 0.001 in elicitation stage, lr = 0.01 in BO stage
	\item Scheduler: CosineAnnealingLR with $T_{max} = 20$ and $eta_{min}=0.0001$ in elicitation stage, no scheduler in BO.
	\item Batch size: 10 for preference data, 5 for regression data
	\item Number of epochs: 100 in elicitation stage, 200 in BO stage
	\item Bayesian linear layer: 3 shared hidden layers with weight prior $\mathcal{N}(0, 0.1)$, width [100, 30, 15]
	\item Activation function: Tanh
\end{itemize}

\section{Additional experiments} \label{appC}
We further investigate the performance of our expert-augmented BO with different elicitation budgets. We use the same objective functions as in Section~\ref{sec:exp_sim} and simulate four different levels of the experts. We use the same configurations as in the previous experiments, the details can be found in Section~\ref{sec:bo_setting}. 

\begin{figure}[t]
  \centering
  \centerline{\includegraphics[width=0.85\textwidth]{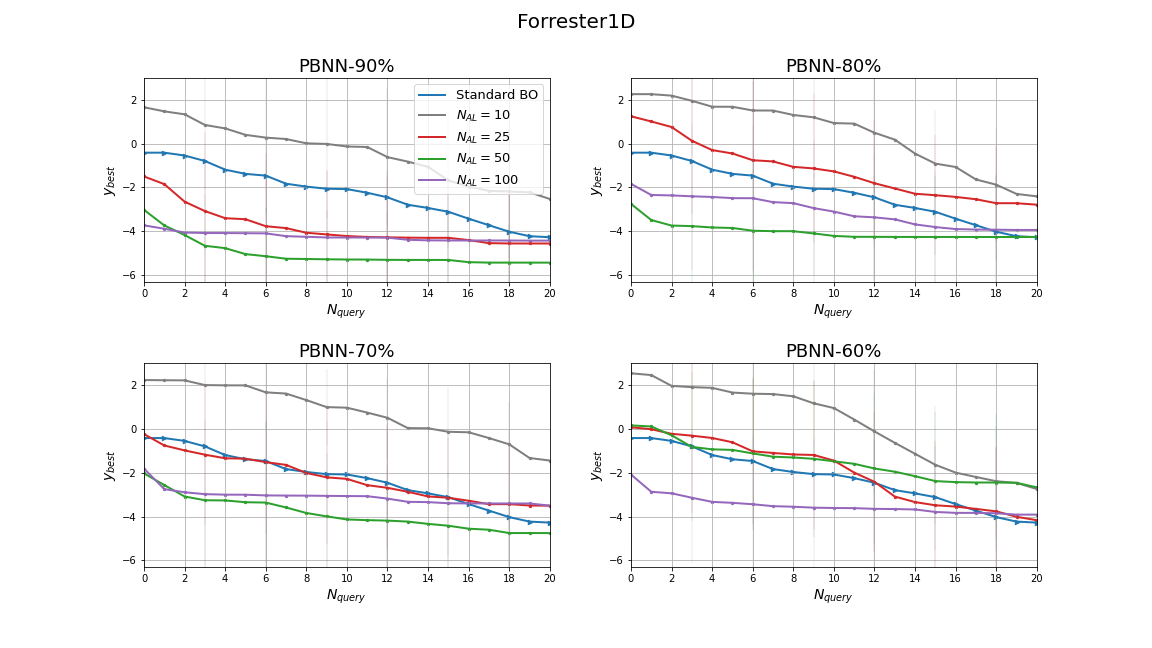}}
  \caption{A BO comparison on "Forrester1D" function with different knowledge elicitation budget. We simulate 4 experts with different levels of knowledge, in each subplot we use the same level of the expert. The results are averaged over 20 simulations.}
  \label{fig:forrester_extra}
\end{figure}

\begin{figure}[t]
  \centering
  \centerline{\includegraphics[width=0.85\textwidth]{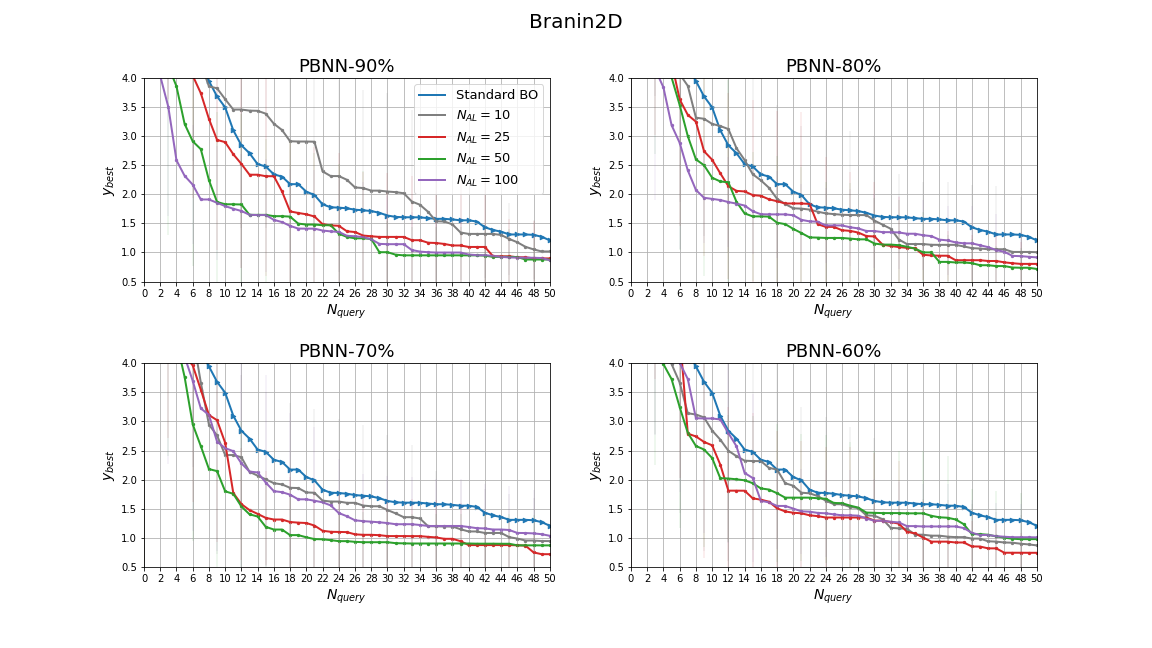}}
  \caption{A BO comparison on "Branin2D" function with different knowledge elicitation budget. We simulate 4 experts with different levels of knowledge, in each subplot we use the same level of the expert. The results are averaged over 20 simulations.}
  \label{fig:branin_extra}
\end{figure}

\begin{figure}[t]
  \centering
  \centerline{\includegraphics[width=0.85\textwidth]{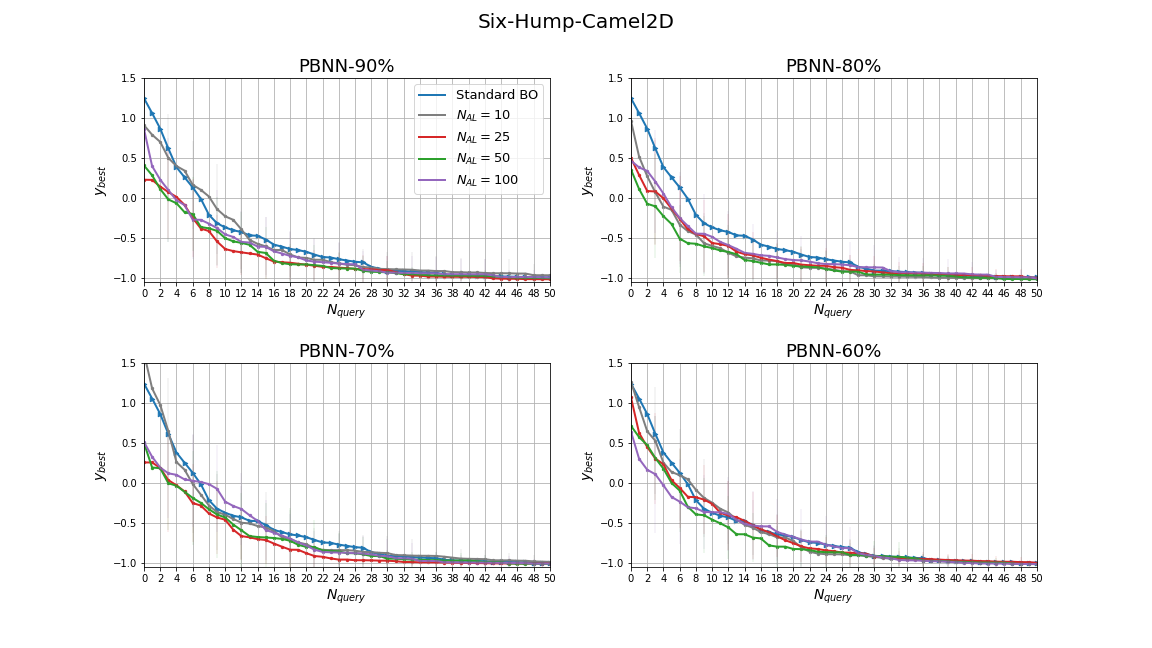}}
  \caption{A BO comparison on "Six-Hump-Camel2D" function with different knowledge elicitation budget. We simulate 4 experts with different levels of knowledge, in each subplot we use the same level of the expert. The results are averaged over 20 simulations.}
  \label{fig:six_hump_camel_extra}
\end{figure}

\begin{figure}[t]
  \centering
  \centerline{\includegraphics[width=0.85\textwidth]{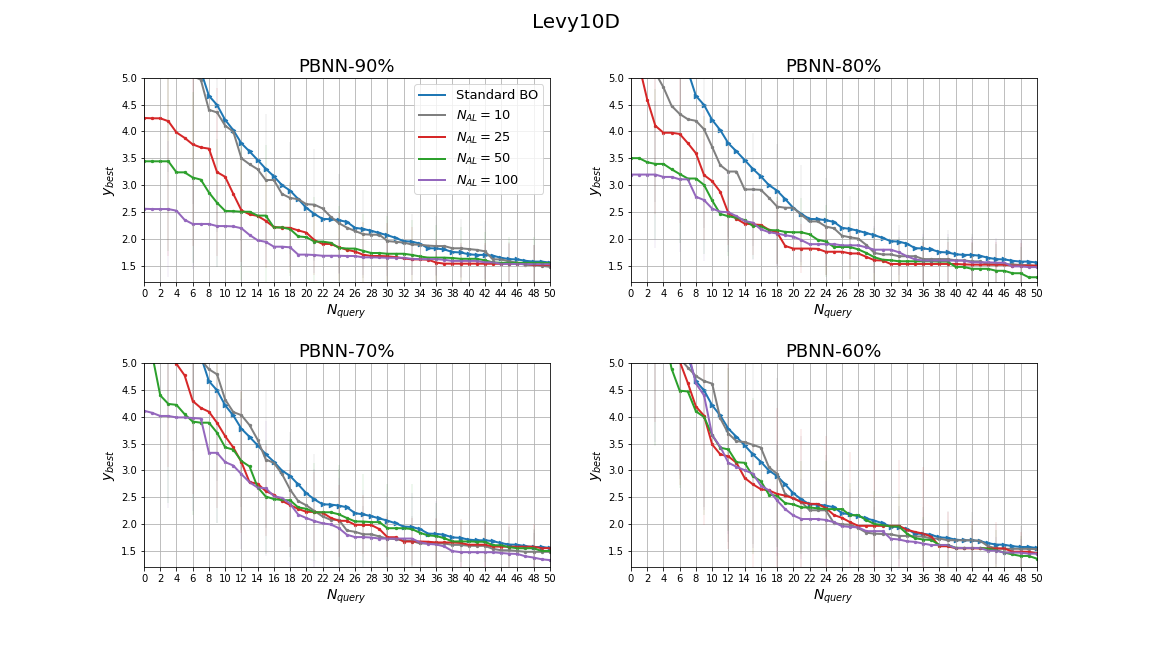}}
  \caption{A BO comparison on "Levy10D" function with different knowledge elicitation budget. We simulate 4 experts with different levels of knowledge, in each subplot we use the same level of the expert. The results are averaged over 20 simulations.}
  \label{fig:levy_extra}
\end{figure}

The results are shown in Figures~\ref{fig:forrester_extra} \ref{fig:branin_extra}, \ref{fig:six_hump_camel_extra}, \ref{fig:levy_extra}. The overall performance behaves as expected. With a larger elicitation budget, the acceleration of BO is more obvious. We notice that the performance is even worse under a very limited budget than standard BO, i.e., $N_{AL}=10$. We guess the reason behind this situation is that the insufficient preference training data makes the network prone to overfitting, hence misguiding the training of the surrogate model during the MTL stage. Moreover, in some figures, we can see the performance between $N_{AL}=50$ and $N_{AL}=100$ is very close, which implies that there is no need to over-query for the expert under some relatively easy-to-optimize functions since the expert knowledge will then dominate the actual BO regression data and slow down the optimization. In this case, we should consider lowering the value of $\alpha$ (Equation~\ref{eq:combined}).

\end{document}